\documentclass[journal]{IEEEtran}
\usepackage{hyperref}
\usepackage{wrapfig}
\usepackage{cite}
\usepackage{algorithm}
\usepackage{booktabs}
\usepackage{multirow}
\usepackage{ulem}

\usepackage{amsmath,amsfonts}
\usepackage{algorithmic}
\usepackage{array}
\usepackage[caption=false,font=normalsize,labelfont=sf,textfont=sf]{subfig}
\usepackage{textcomp}
\usepackage{stfloats}
\usepackage{url}
\usepackage{verbatim}
\usepackage{graphicx}

\begin{document}

\title{Segment Anything Model is a Good Teacher for Local Feature Learning}

\author{Jingqian Wu, \and
Rongtao Xu, \and
Zach Wood-Doughty, \and
Changwei Wang, \and \\
Shibiao Xu,~\IEEEmembership{Member,~IEEE,} \and
Edmund Y. Lam,~\IEEEmembership{Fellow,~IEEE}

\thanks{Jingqian Wu and Edmund Y. Lam are with The University of Hong Kong, Pokfulam.}
\thanks{Rongtao Xu is with the State Key Laboratory of Multimodal Artificial Intelligence Systems, Institute of Automation, Chinese Academy of Sciences, Beijing, China and School of Artificial Intelligence, University of Chinese Academy of Sciences, Beijing 100190, China.}%
\thanks{Zach Wood-Doughty is with Northwestern University, Evanston, IL 60201, USA}%
\thanks{Changwei Wang (Corresponding author) is with the Key Laboratory of Computing Power Network and Information Security, Ministry of Education, Shandong Computer Science Center (National Supercomputer Center in Jinan), Qilu University of Technology (Shandong Academy of Sciences), Jinan, 250013, China; Shandong Provincial Key Laboratory of Computer Networks, Shandong Fundamental Research Center for Computer Science, Jinan, China; the State Key Laboratory of Multimodal Artificial Intelligence Systems, Institute of Automation, Chinese Academy of Sciences, Beijing, China. Email: wangchangwei2019@ia.ac.cn.}
\thanks{Shibiao Xu is with the School of Artificial Intelligence, Beijing University of Posts and Telecommunications, Beijing 100876, China.}
}

\markboth{IEEE TRANSACTIONS ON IMAGE PROCESSING,~Vol.~xx, No.~x, x~2024}%
{Shell \MakeLowercase{\textit{et al.}}: Semantic Enhanced Deep Local Features}

\maketitle

\begin{abstract}
Local feature detection and description play an important role in many computer vision tasks, which are designed to detect and describe keypoints in any scene and any downstream task.
Data-driven local feature learning methods need to rely on pixel-level correspondence for training. However, a vast number of existing approaches ignored the semantic information on which humans rely to describe image pixels.
In addition, it is not feasible to enhance generic scene keypoints detection and description simply by using traditional common semantic segmentation models because they can only recognize a limited number of coarse-grained object classes.
In this paper, we propose SAMFeat to introduce SAM (segment anything model), a foundation model trained on 11 million images, as a teacher to guide local feature learning. SAMFeat learns additional semantic information brought by SAM and thus is inspired by higher performance even with limited training samples.
To do so, first, we construct an auxiliary task of {Attention-weighted Semantic Relation Distillation} (ASRD), which adaptively distillates feature relations with category-agnostic semantic information learned by the SAM encoder into a local feature learning network, to improve local feature description using semantic discrimination.
Second, we develop a technique called Weakly Supervised Contrastive Learning Based on Semantic Grouping (WSC), which utilizes semantic groupings derived from SAM as weakly supervised signals, to optimize the metric space of local descriptors.
Third, we design an Edge Attention Guidance (EAG) to further improve the accuracy of local feature detection and description by prompting the network to pay more attention to the edge region guided by SAM.
SAMFeat's performance on various tasks such as image matching on HPatches, and long-term visual localization on Aachen Day-Night showcases its superiority over previous local features.
The release code is available at~\href{https://github.com/vignywang/SAMFeat}{https://github.com/vignywang/SAMFeat}.
\end{abstract}

\IEEEpeerreviewmaketitle

\section{Introduction}
\label{sec:intro}
\IEEEPARstart{L}{ocal} feature detection and description is a basic task of computer vision, which is widely used in image matching~\cite{hpatches}, structure from motion~(SfM)~\cite{colmap}, simultaneous localization and mapping~(SLAM)~\cite{orb-slam2}, visual localization~\cite{benchmark1}, and image retrieval~\cite{rank} tasks.
Traditional schemes such as SIFT~\cite{sift}, and ORB~\cite{rublee2011orb} based hand-crafted heuristics are not able to cope with drastic illumination and viewpoint changes~\cite{hpatches}.
Under the wave of deep learning, many data-driven local feature learning methods~\cite{detone2018superpoint,tyszkiewicz2020disk} have recently achieved excellent performance.
While many works have been done for training local descriptors based on completely accurate and dense ground truth correspondences~\cite{li2018megadepth} between image pairs, a vast number of these works ignored the semantic information on which humans rely to describe image pixels.
Even though few previous works adapted the straightforward idea of using traditional common semantic segmentation models to facilitate the detection and description of keypoints, it is not feasible in practice because they can only recognize a limited number of coarse-grained object categories and are not competent for keypoint detection and description in generalized scenarios~\cite{kirillov2023segment}.

Recently, foundation models~\cite{bommasani2021opportunities} have revolutionized the field of artificial intelligence. These models, trained on billions of examples, presented strong zero-shot generalization capabilities across a variety of downstream tasks.
In this study, we advocate the integration of SAM~\cite{kirillov2023segment}, a foundation model that is able to segment "anything" in "any scene", into the realm of local feature learning. This synergy enhances the robustness and enriches the supervised signals available for local feature learning, encompassing high-level category-agnostic semantics and detailed low-level edge structure information.

\begin{figure*}[h]
\begin{center}
  \includegraphics[width=0.8 \linewidth]{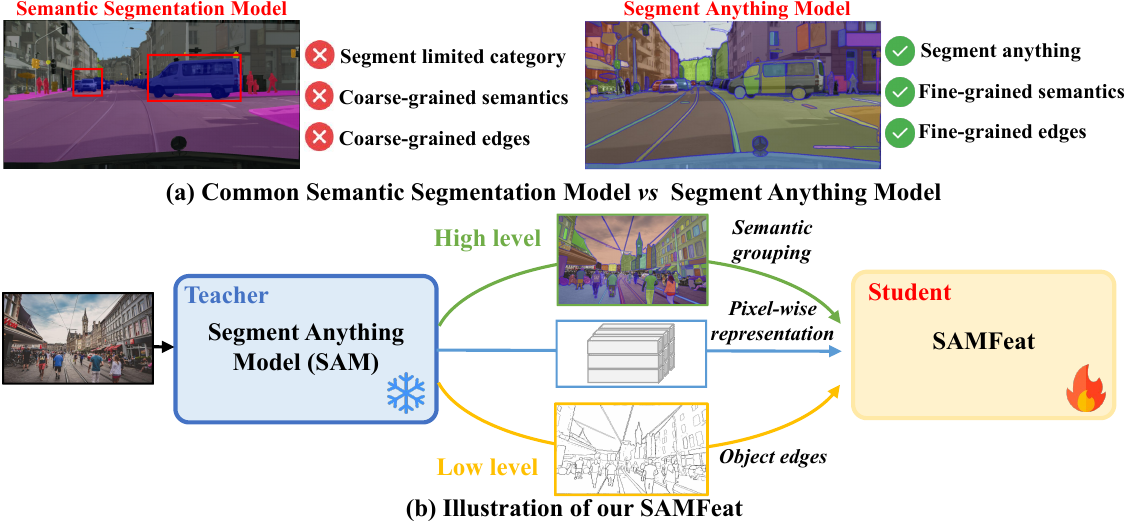}
\end{center}
   \caption{(a): Difference between segment anything model and common semantic segmentation model. (b): Schematic diagram of proposed SAMFeat.}
\label{fig:intro}
\end{figure*}

In recent years, works have attempted to introduce pixel-level semantics of images (\textit{i.e.} semantic segmentation) into local feature learning-based visual localization.
Some methods utilized semantic information to filter keypoints~\cite{xue2022efficient} and optimize matching~\cite{schonberger2018semantic}, while other works utilized semantic information~\cite{xue2023sfd2} to guide the learning of keypoints detection and improve the performance of the local descriptors in a specific visual localization setting by using feature-level distillation.
However, these visual localization pipelines and methods are based on common semantic segmentation models and are difficult to generalize to feature matching tasks. As shown in Fig.~\ref{fig:intro} (a), this is because, on one hand, common semantic segmentation can only assign semantics to limited categories (\textit{e.g.} cars, streets, people) which is difficult to generalize to generic scenarios and open world situations~\cite{kirillov2023segment}.
On the other hand, the semantic information for semantic segmentation is coarse-grained, \textit{e.g.}, pixels of wheels and windows are given the same labels for a car. This is detrimental to mining the unique discriminative properties of local features.




The recent SAM~\cite{kirillov2023segment} is a visual foundation model trained on 11 million images that can segment any objects based on prompt input.
Compared to common semantic segmentation models, SAM has three unique properties that can be used to fuel local feature learning.
\textbf{\textit{i)}}
SAM is trained on a large amount of data, and therefore, can segment any object and can be adapted to any scene rather than being limited to certain categories and scenes like common semantic segmentation models. SAM's robust zero-shot performance could be a helpful enhancement in learning and describing features in complex scenes.
\textbf{\textit{ii)}}
SAM can obtain fine-grained component-level semantic segmentation results, thus allowing for more accurate modeling of semantic relationships between pixels. In addition, SAM can derive fine-grained category-agnostic semantic masks that can be used as semantic groupings of pixels to guide local feature learning.
\textbf{\textit{iii)}}
SAM can detect more detailed edges, whereas edge regions tend to be more prone to critical points and contain more distinguishing information, which helps feature learning by providing accurate guidance in keypoint localization.

In our SAMFeat, we propose three special strategies to boost the performance of local feature learning based on these three properties of SAM.
\textbf{First,} we construct an auxiliary task of {Attention-weighted Semantic Relation Distillation} (ASRD) for distilling category-agnostic pixel semantic relations learned by the SAM encoder into a local feature learning network with attention guide, thus using semantic discriminative to improve local feature description.
\textbf{Second,} we develop a technique called Weakly Supervised Contrastive Learning Based on Semantic Grouping (WSC) to optimize the metric space of local descriptors using SAM-derived semantic groupings as weakly supervised signals.
\textbf{Third,} we design an Edge Attention Guidance (EAG) to further improve the localization accuracy and description ability of local features by prompting the network to pay more attention to the edge region.
Since the SAM model is only used as a teacher during training, our SAMFeat can efficiently extract local features during inference without burdening the computational consumption of the SAM network.

\section{Related Work}
\noindent\textbf{Local Features and Beyond.} 
Early hand-crafted local features have been investigated for decades and are comprehensively evaluated in~\cite{traditional_evaluation2}.
In the wave of deep learning, many data-driven learnable local features have been proposed for improving detectors based on different focuses on~\cite{affnet,keynet}, descriptors~\cite{l2net,hardnet,sosnet,luo2019contextdesc}, and end-to-end detection and description~\cite{lift,lfnet,superpoint,r2d2,d2net,aslfeat,wang2022mtldesc}.
Beyond localized features, some learnable advanced matchers have recently been developed to replace the traditional nearest neighbor matcher (NN) to get more accurate matching results.
Sparse matchers such as SuperGlue~\cite{sarlin2020superglue} and LightGlue~\cite{lindenberger2023lightglue} take off-the-shelf local features as input to predict matches using a GNN or Transformer, however, their time complexity scales quadratically with the number of keypoints.
Dense matchers~\cite{sun2021loftr,yu2023adaptive} compute the correspondence between pixels end-to-end based on the correlation volume, while they spend more memory and space consumption than sparse matchers~\cite{xue2023sfd2}.
Our work centers on enhancing the efficiency and performance of an end-to-end generalized local feature learning approach. We aim to achieve performance comparable to advanced matchers while only using nearest-neighbor matching across various downstream tasks. This is particularly crucial in resource-constrained scenarios demanding high operational efficiency.

\noindent\textbf{Segment Anything Model.} 
The Segment Anything Model (SAM)~\cite{kirillov2023segment} has achieved remarkable advancements in expanding the scope of segmentation tasks, thereby significantly fostering the evolution of fundamental models in computer vision.
SAM incorporates prompt learning techniques in the field of NLP to flexibly implement model building and builds an image engine through interactive annotations, which performs better in techniques such as instance analysis, edge detection, object proposal, and text-to-mask.
SAM is specifically designed to address the challenge of segmenting a wide range of objects in complex visual scenes. Unlike traditional approaches that focus on segmenting specific object classes, SAM's primary objective is to segment anything, providing a versatile solution for diverse and challenging scenarios. Many works \cite{he2023scalable, kristan2021ninth} now build upon SAM for downstream vision tasks such as medical imaging, video, data annotation, \textit{etc}~\cite{zhang2023comprehensive}. 
Unlike them, we advocate for the application of SAM to local feature learning.
To the best of our knowledge, our work is the first to apply SAM to feature learning and matching tasks.
There is, indeed, other newly proposed state-of-the-art work that incorporates other visual foundation models to tackle feature learning tasks. For example, ROMA \cite{edstedt2023roma} proposed to incorporate the encoder from DINO-V2 \cite{oquab2023dinov2} directly into their feature learning framework and fine-tune it in the training stage.
However, the computational cost for training and inference of such a method is extremely high.
Since there are needs for high operational efficiency requirements in real-time local feature matching applications, we choose not to incorporate SAM directly into the pipeline. To leverage the knowledge from SAM while maintaining high efficiency and inference speed, we treat SAM as a teacher to bootstrap local feature learning, thus using SAM only in the training phase.

\noindent\textbf{Semantics in Local Feature Learning.}
Prior to our work, semantic information had solely been incorporated into the visual localization task as a means to mitigate the challenges posed by low-level local features when dealing with severe image variations.
Some early works incorporated semantic segmentation into the visual localization pipeline for filtering matching points~\cite{huang2021vs,hu2020dasgil}, improving 2D-3D matching~\cite{toft2018semantic,shi2020dense}, and estimating camera position~\cite{toft2017long}.
Some recent works~\cite{fan2022learning,xue2023sfd2} have attempted to introduce semantics into local feature learning to improve the performance of visual localization.
Based on the assumption that high-level semantics are insensitive to photometric and geometric, they enhance the robustness of local descriptors on semantic categories by distilling features or outputs from semantic segmentation networks.
However, semantic segmentation tasks can only segment certain specific categories (\textit{e.g.}, visual localization-related street scenes), preventing such approaches from generalizing to open-world scenarios and making them effective only on visual localization tasks.
In contrast, we introduce SAM for segmenting any scene as a distillation object and propose the category-agnostic Attention-weighted Semantic Relation Distillation (ASRD) scheme to enable local feature learning to enjoy semantic information in scenes beyond visual localization.
In addition, we also propose Weakly Supervised Contrastive Learning Based on Semantic Grouping (WSC) and Edge Attention Guidance (EAG) to further motivate the performance of local features based on the special properties of SAM.
Based on the above improvements, our SAMFeat makes it possible for local feature learning to more fully utilize semantic information and benefit in a wider range of scenarios.

\section{Methodology}
\label{sec:Method}

\subsection{Overview}
\begin{figure}[t]
	\centering
	\includegraphics[width=\linewidth,scale=1.00]{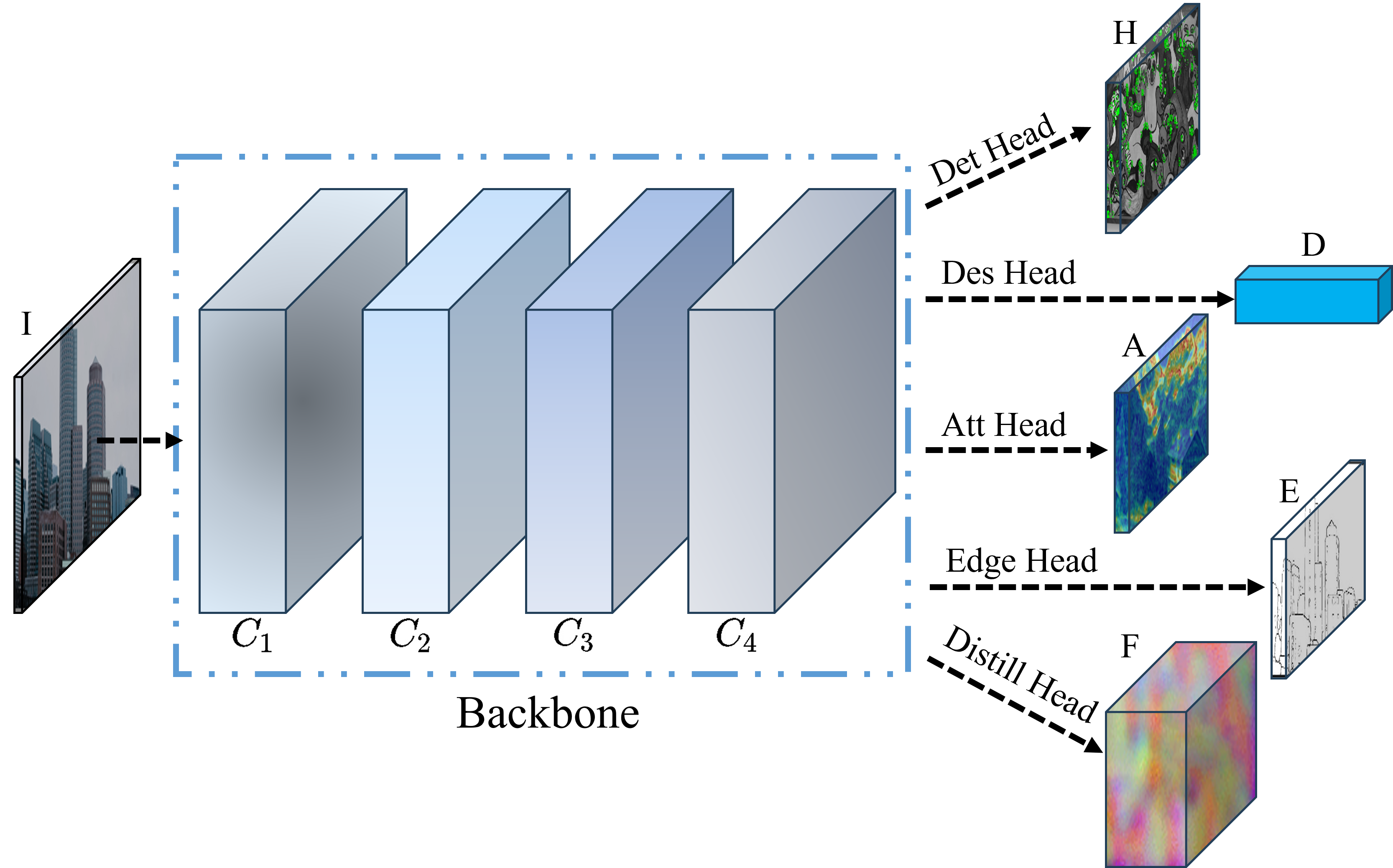}
	\caption{The overview of our SAMFeat, which performs feature detection, description, edge depiction, and feature distillation end-to-end.}
	\label{fig:simple overview}
\end{figure}

Our SAMFeat uses a backbone for feature extraction, along with several heads serving different purposes end-to-end. The simple overview of SAMFeat is shown in Figure \ref{fig:simple overview} and the detailed network structure is shown in Figure~\ref{fig:method}.

\noindent\textbf{Backbone.}
We use an eight-layer VGG-style backbone encoder following \cite{detone2018superpoint} to extract feature maps. The encoder consists of $3 \times 3$ convolutional layers, relu layers, and max-pooling layers.
For a $H\times W$ image $I$, we concatenate multiscale feature map outputs ($C_{1} \in \mathbb{R}^{H\times W \times 64},C_{2} \in \mathbb{R}^{\frac{1}{2}H\times \frac{1}{2}W \times 64},C_{3} \in \mathbb{R}^{\frac{1}{4}H\times \frac{1}{4}W \times 128},C_{4} \in \mathbb{R}^{\frac{1}{8}H\times \frac{1}{8}W \times 128})$ delivered to the keypoint detection head (Det Head), edge head (Edge Head), attention head (Att Head), distillation head (Distall Head), and descriptor head (Des Head).

\noindent\textbf{Keypoint Detection Head.}
We employ four detection layers to predict keypoint heatmaps at various scales. To integrate these predictions, we upsample the heatmaps to match the image dimensions of $h \times w$. We then use four learnable weights to merge the heatmaps from different scales, thereby predicting the final keypoints and calculating the associated loss. Each detection head receives direct supervision through the detector loss, providing a form of deep supervision as described in \cite{wang2022mtldesc}.

We utilize a weighted binary cross-entropy loss for the detector due to the significant imbalance between keypoints and non-keypoints. With the predicted keypoint heatmap \( H \in \mathbb{R}^{h \times w} \) and the pseudo-ground truth label \( G \in \mathbb{R}^{h \times w} \), the detector loss $\mathcal{L}_{\text{det}}$ is defined as follows:



\begin{align}
\mathcal{L}_{\text{bce}}(h, g) &= -\lambda g \log(h) - (1 - g) \log(1 - h), \\
\mathcal{L}_{\text{det}} &= \frac{1}{hw} \sum_{u,v} \mathcal{L}_{\text{bce}}(H_{u,v}, G_{u,v}),
\end{align}

where the weight $\lambda$ is empirically set to 200.

\noindent\textbf{Attention Head.}
\label{Att Head}
The Attention Head is designed to generate the attention map $A \in \mathbb{R}^{\frac{1}{4} h \times \frac{1}{4} w}$.
Specifically, we obtain \( C_{\text{cat}} \) by concatenating the feature maps \( C_1, C_2, C_3, \) and \( C_4 \) from the backbone network. Next, \( C_{\text{cat}} \) is averaged across the channel dimension, and an attention map \( A \) is generated from this averaged feature map using a 3 × 3 convolutional layer followed by a softplus activation function.

The attention map proves to be highly useful in descriptor optimization, distillation optimization, and matching processes \cite{wang2023attention}. In terms of descriptor and distillation optimization, we provide a detailed analyze of motivation in Section \ref{Description Head} and Section \ref{SAMFeat}. For matching, attention-weighted local descriptors are more effective. Regions with high attention scores in one image are likely to match with similar high-score regions in another image, reducing the matching space and enhancing accuracy. These consistent attention scores serve as prior information, making local descriptor matching more efficient and reliable. The Attention Head will be jointly optimized during the optimization process of both descriptor generation and feature distillation.

\noindent\textbf{Feature Description Head.}
\label{Description Head}
Initially, we densely extract descriptors for all pixels to form the set \(D\). We then extract descriptors \(d\) for individual pixels at their respective locations. In accordance with previous studies \cite{wang2022mtldesc}, we utilize L2 normalization to obtain the local descriptors, defined as:

\begin{equation}
d = \frac{D_{(i,j)}}{\| D_{(i,j)} \|},
\label{descriptor function}
\end{equation}

where \(\| \cdot \|\) represents the L2 norm and \((i, j)\) denotes the index position of the local descriptor \(d\).

Inspired by \cite{wang2023attention, wang2022mtldesc}, the descriptor loss splits the optimization into two parts: the descriptor's angle and the consistent attention weight. This differs from the standard triplet loss, which only focuses on the angle component. Positive samples have converging attention scores, while negative samples diverge, leading to a consistent distribution of attention scores across image pairs. Higher attention scores significantly influence the gradient, allowing the network to prioritize more relevant samples and avoid optimizing descriptors for less informative pixels, like the sky or grass.

The descriptor Loss is formulated to jointly optimize local descriptors and consistent attention. Building on previous research \cite{wang2022mtldesc, wang2023attention}, we use the corresponding point sets \( (P, P') \) obtained from ground-truth camera parameters and depths to supervise the training of local descriptors. For an image pair \( (I, I') \), dense descriptors \( D, D' \) and attention maps \( A, A' \) are extracted using our SAMFeat method.

Given a point set \( P \) of size \( M \) in image \( I \) and the corresponding points \( P' \) in image \( I' \), the local descriptors of \( P, P' \) are denoted by Equation \ref{descriptor function} as \( d_i \) respectively, where \( i \in \{1, \ldots, M\} \). The corresponding score of the descriptor \( d_i \) on the attention map \( A \) is denoted as \( \nu_i \). Thus, the attention-weighted descriptor is defined as \( y_i = \nu_i \cdot d_i \). For \( y_i \), its positive distance \( \| y_i \|^{+} \) is defined as:

\begin{equation}
\| y_i \|^{+} = \| \nu_i \cdot d_i - \nu'_i \cdot d'_i \|_2,
\end{equation}

and its hardest negative distance \( \| y_i \|^{-} \) is defined as:

\begin{equation}
\| y_i \|^{-} = \min_{j \in \{1, \ldots, M\}, j \neq i} \| \nu_i \cdot d_i - \nu'_j \cdot d'_j \|_2.
\end{equation}

The overall descriptor loss is the sum of the individual losses:

\begin{equation}
\mathcal{L}_{\text{det}}(y) = \sum_{i=1}^{N} \frac{e^{\nu/T}}{\sum_{j=1}^{M} e^{\nu_j/T}} \max(0, \| y_i \|^{+} - \| y_i \|^{-} + 1),
\end{equation}

where \( \nu \) is the attention score corresponding to \( y \), and \( T \) is a smoothing factor that adjusts the effect of attention weighting on the loss. \( T \) is set to $15$ following \cite{wang2023attention}.

\noindent\textbf{Edge Head.}
The Edge Head is designed to learn edges, corners, and keypoints information from SAM.
The Edge Map $E$ is generated simply via a convolutional and sigmoid layer, taking the concatenated features outputted from the shared backbone. With a simple edge map supervision loss, SAMFeat is able to mimic and generate accurate edge maps learned from SAM edge knowledge as shown in Fig \ref{fig:Edge_Learning}.
To further utilize the learned edge information, we designed the Edge Attention Guidance Module that enhanced keypoint detection.
The corresponding learning loss for Edge Map and the mechanism behind the guidance module will be introduced with details in Section \ref{Edge Attention Guidance.}

\noindent\textbf{Distill Head.}
The Distill Head is designed to distillate the robust prior feature knowledge from the powerful SAM backbone for better image and scene understanding and recognition. The head is constructed by one convolutional operation, and the learned prior feature will also be fused into the network by another convolutional operation. Supervision loss and feature fusion process will be presented with details in Section \ref{Distill Head.}.

\subsection{Gifts from SAM}
\label{sec:3.2}
SAM~\cite{kirillov2023segment} is a newly released visual foundation model for segmenting any objects and has strong zero-shoot generalization due to the fact that it is trained using 11 million images and 1.1 billion masks. Due to its scale, model distillation~\cite{hinton2015distilling} is deployed in this work. We freeze the weights of SAM and use its output as pseudo-ground truth to guide more accurate and robust local feature learning. In this subsection, we introduce how we utilize three gifts from SAM to enhance our SAMFeat.
Shown in Figure~\ref{fig:method}, we input the image $I$ into the SAM~\cite{kirillov2023segment} with frozen parameters and then simply processed to produce the following three outputs for guided local feature learning.

\textbf{Pixel-wise Representations Relationship}: SAM's image encoder trained from 11 million images is used to extract image representations for assigning semantic labels.
The representation of the encoder outputs implies a valuable semantic correspondence, \textit{i.e.}, pixels of the same semantic object are closer together.
To eliminate the effect of specific semantic categories on generalizability, we adopt relations between representations as distillation targets.
SAM's encoder outputs $\mathcal{F} \in \mathbb{R}^{\frac{1}{8}H\frac{1}{8}W\times C}$, where $C$ is the channel number for feature map. The pixel-wise representations relationship can be defined as $\mathcal{R} \in \mathbb{R}^{\frac{1}{8}H\frac{1}{8}W\times \frac{1}{8}H\frac{1}{8}W}$, where $\mathcal{R}(i,j)= \frac{\mathcal{F}(i) \cdot \mathcal{F}(j)}{|\mathcal{F}(i)||\mathcal{F}(j)|}$.

\textbf{Semantic Grouping}: 
We use the automatically generating masks function\footnote{https://github.com/facebookresearch/segment-anything/} of SAM to obtain fine-grained semantic groupings. Specifically, it works by sampling single-point input prompts in a grid over the image, and SAM can predict multiple masks from each of them. Then, masks are filtered for quality and deduplicated using non-maximal suppression \cite{kirillov2023segment}.
The semantic grouping of the output can be defined as ${G} \in \mathbb{R}^{H\times W \times N}$, where $N$ is the number of semantic groupings. Notice that semantic grouping differs from semantic segmentation in that each grouping does not correspond to a specific semantic category (\textit{e.g.} buildings, car, and person).

\textbf{Edge Map}: The binary edge map ${E} \in \mathbb{R}^{H \times W \times 1}$ is derived directly~\footnote{https://github.com/ymgw55/segment-anything-edge-detection} from the segmentation results of SAM, which highlights the fine-grained object boundaries.

\subsection{SAMFeat}
\begin{figure*}[t]
\setlength{\abovecaptionskip}{0.cm} 
\setlength{\belowcaptionskip}{-0.6cm}
\begin{center}
  \includegraphics[width=0.9 \linewidth]{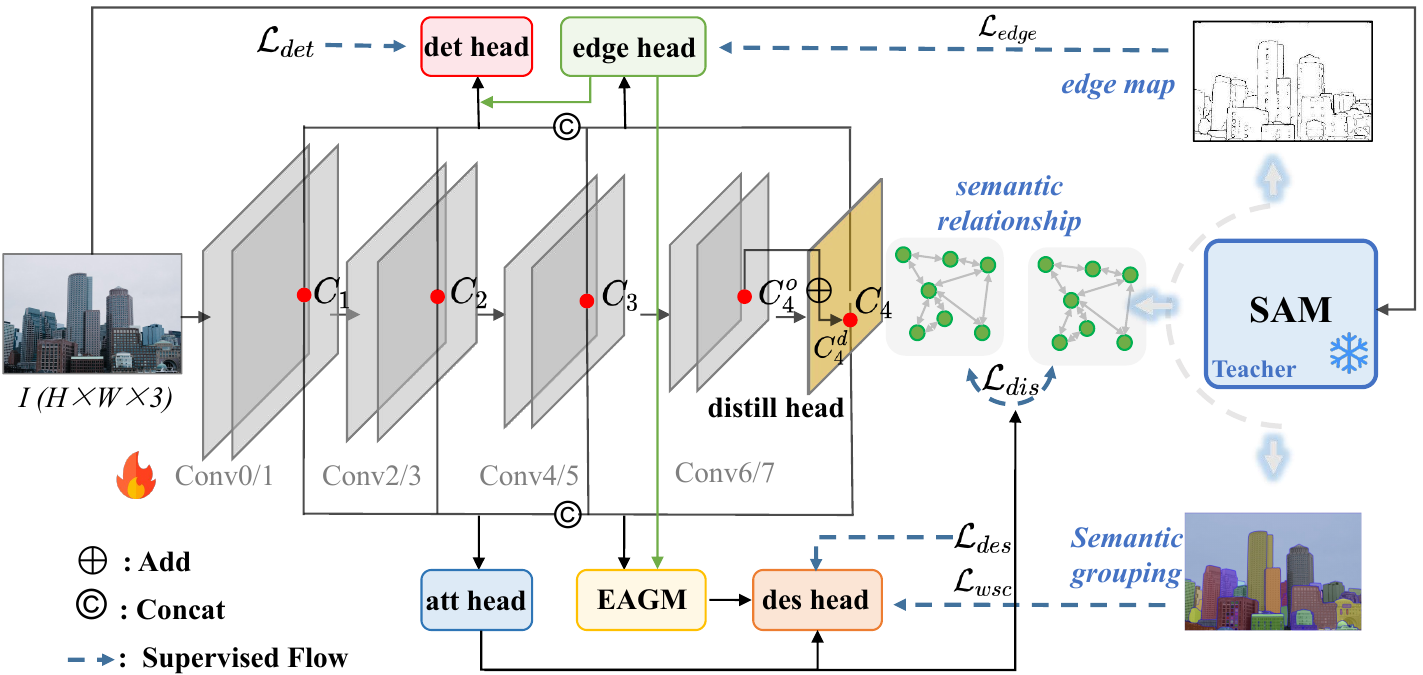}
\end{center}
   \caption{The detailed overview of SAMFeat. Notice that SAM is only applied in the training phase, while there is no computational cost in the inference phase.
   }

\label{fig:method}
\end{figure*}

\label{SAMFeat}
Thanks to the gifts of the foundation model, SAM, we are able to consider SAM as a knowledgeable teacher with intermediate products and outputs to guide the learning of local features.
First, we employ Attention-weighted Semantic Relation Distillation (ASRD) to distill the category-agnostic semantic relations in the SAM encoder into SAMFeat, thereby enhancing the expressive power of local features by introducing semantic distinctiveness.
Second, we utilize the high-level semantic grouping of SAM outputs to construct Weakly Supervised Contrastive Learning Based on Semantic Grouping (WCS), which provides cheap and valuable supervision for local descriptor learning.
Third, we design an Edge Attention Guidance (EAG) to utilize the low-level edge structure discovered by SAM to guide the network to pay more attention to these edge regions, which are more likely to be detected as keypoints and rich in discriminative information during local feature detection and description.

\noindent\textbf{Attention-weighted Semantic Relation Distillation.} 
\label{Distill Head.}
SAM aims to obtain the corresponding semantic masks based on the prompt, so the encoder output representation of SAM is rich in semantic discriminative information.
Unlike semantic segmentation, SAM does not project pixels to a specified semantic category, so we resort to distilling the semantics contained in the encoder by exploiting the relative relationship between pixels (\textit{i.e.}, pixel representations of the same object are closer together).

However, not every pixel is equally important for local feature learning, and forcing the network to learn a large proportion of background pixels (\textit{e.g.}, the sky) can hinder the learning of discriminative foreground regions. We therefore advocate a greater focus on discriminative foreground regions in the distillation process.
In contrast to classical relational distillation~\cite{park2019relational}, we propose Attention-weighted Semantic Relation Distillation (ASRD) to guide the distillation process to focus on valuable relational pairs to further motivate the transfer of knowledge that facilitates local feature matching.

\begin{figure}[htbp]
	\centering
	\includegraphics[width=\linewidth,scale=1.00]{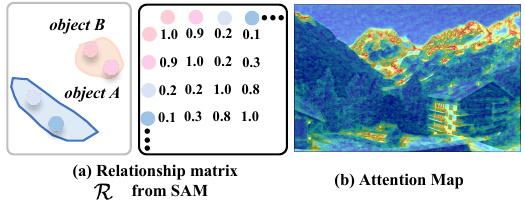}
	\caption{Schematic diagrams of Relationship matrix and Attention map.}
	\label{fig:ASRD}
\end{figure}

Specifically, $C^{o}_{4}$ is exported from \textit{Conv7} layer and then imported into the distillation head to get $C^{d}_{4} \in \mathbb{R}^{\frac{1}{8}H\times \frac{1}{8}W \times 128}$.  
Following the operations reported in Sec.~\ref{sec:3.2}, the semantic relation matrix of $C^{d}_{4}$ can be defined as $\mathcal{R}^{'}$.
As shown in Figure~\ref{fig:ASRD}, we distill the semantic relation matrix by imposing L1 loss in order to obtain semantic discriminativeness for $C^{d}_{4}$.
$\mathcal{R}^{'}$ and $\mathcal{R}$ are the corresponding student (SAMFeat) and teacher (SAM) relation matrix. 
As shown in Fig.~\ref{fig:ASRD} (b), we use the attention map obtained in Section~\ref{Att Head} to construct the weight matrix used to weight the relational distillations.
We flatten the attention map into a 1-dimensional vector $V_{A}$, and then multiply $V_{A}$ and the transpose $V_{A}^{T}$ of $V_{A}$ by matrix multiplication to obtain the weight matrix $\mathcal{W}$, which can be formally defined as follows:
\begin{equation}
    \mathcal{W} =  V_{A} \times  V_{A}^{T},
\end{equation}
where the elements of $\mathcal{W} \in \mathbb{R}^{\frac{1}{8}H\times \frac{1}{8}W}$ and $\mathcal{R}$ correspond to each other.
Attention-weighted Semantic Relation Distillation Loss $\mathcal{L}_{dis}$ can be defined as:
\begin{equation}
    \mathcal{L}_{dis} =  \frac{\sum_{i,j}^{(\frac{1}{8}H\times \frac{1}{8}W),(\frac{1}{8}H\times \frac{1}{8}W)}|\mathcal{R}_{i,j} - \mathcal{R}^{'}_{i,j}|\cdot e^{W(i,j)}}{\sum_{i,j}e^{W(i,j)}},
\end{equation}
where $N$ is the number of matrix elements, \textit{i.e.}, $(\frac{1}{8}H\times \frac{1}{8}W) \times (\frac{1}{8}H\times \frac{1}{8}W)$.
Since ASRD is category-agnostic, it is possible to generalize local feature distillation semantic information to generic scenarios. A detailed pseudo-code is illustrated in algorithm \ref{algorithm_ASRD}. 


\renewcommand{\algorithmicrequire}{\textbf{Input:}}
\renewcommand{\algorithmicensure}{\textbf{Output:}}
\begin{algorithm}
\caption{Attention-weighted Semantic Relation Distillation}\label{algorithm_ASRD}
        \begin{algorithmic}[1] 
            \REQUIRE Image pair $I_{1}, I_{2}$; Attention Map $A$; $H= W=400$; $C=256$; SAMFeat's encoder $E$; SAM's encoder $E'$.
            \ENSURE SAMFeat's Relationship Matrix $\mathcal{R}$; SAM's Relationship Matrix $\mathcal{R}'$.
            \STATE Given $I_{1}, I_{2}$, an encoded image feature $\mathcal{F} \in \mathbb{R}^{\frac{1}{8}H \times \frac{1}{8}W\times C}$ can be obtained via $E$.
            \STATE Given $I_{1}, I_{2}$, an encoded image feature $\mathcal{F'} \in \mathbb{R}^{64\times64\times C}$ can be obtained via $E'$.
            \STATE Downsample $\mathcal{F'}$ to $\mathcal{F'}_{down} \in \mathbb{R}^{\frac{1}{8}H\frac{1}{8}W\times C}$
            \STATE Downsample $\mathcal{A}$ to $\mathcal{A}_{down} \in \mathbb{R}^{\frac{1}{8}H \times \frac{1}{8}W}$, and apply cross multiplication with itself and apply softmax activation function to obtain the Attention Weight $A_w$ $\in \mathbb{R}^{\frac{1}{8}H\frac{1}{8}W\times \frac{1}{8}H\frac{1}{8}W}$
            \STATE Flatten $\mathcal{F}$ and $\mathcal{F'}_{down}$ then calculate mean on $dim = C$ to obtain $\mathcal{F}_{flatten} \in \mathbb{R}^{\frac{1}{8}H\frac{1}{8}W}$ and $\mathcal{F'}_{flatten} \in \mathbb{R}^{\frac{1}{8}H\frac{1}{8}W}$
            \STATE Construct Attention Weighted Relationship Matrix $\mathcal{R} \in \mathbb{R}^{\frac{1}{8}H\frac{1}{8}W\times \frac{1}{8}H\frac{1}{8}W}$ and $\mathcal{R'} \in \mathbb{R}^{\frac{1}{8}H\frac{1}{8}W\times \frac{1}{8}H\frac{1}{8}W}$ from $\mathcal{F}_{flatten}$ and $\mathcal{F'}_{flatten}$ respectively, where $\mathcal{R}(i,j)= \frac{\mathcal{F}(i) \cdot \mathcal{F}(j)}{|\mathcal{F}(i)||\mathcal{F}(j)|} \cdot A_w(i,j)$, and same applied for $\mathcal{R'}$
            \RETURN $\mathcal{L}_{dis} = |\mathcal{R}$ - $\mathcal{R'}|$.  
        \end{algorithmic}
\end{algorithm}

\begin{figure}[]
	\centering
	\includegraphics[width=\linewidth,scale=1.00]{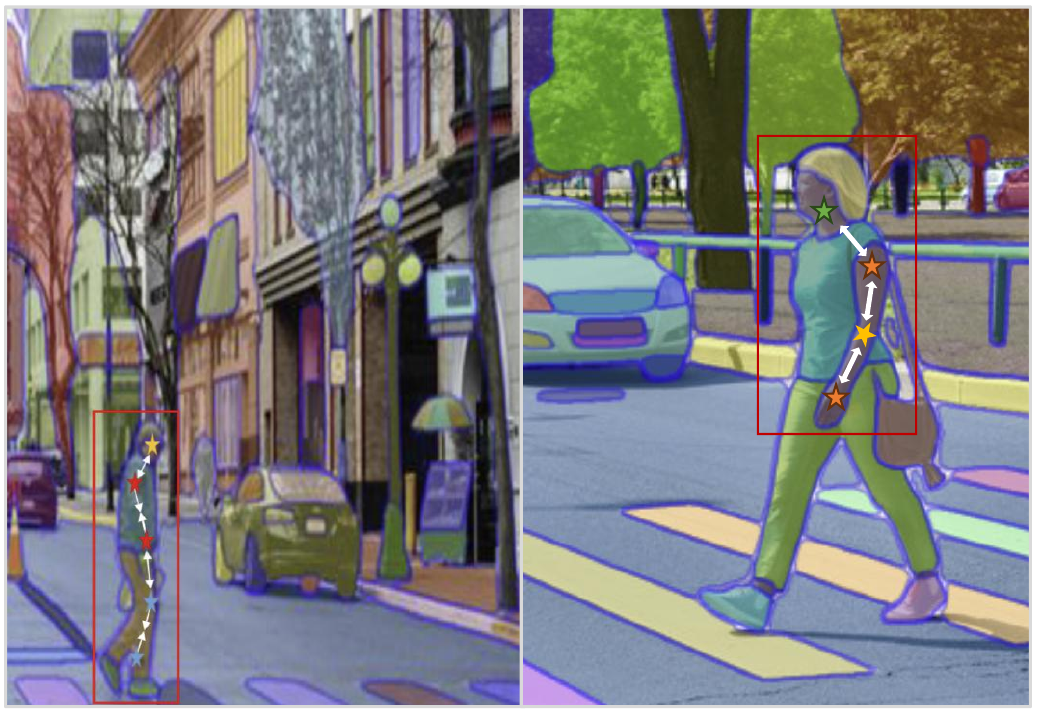}
	\caption{Example of Semantic Grouping. Different colored stars represent sampling points in different semantic groupings.}
	\label{fig:wsg}
\end{figure}

\noindent\textbf{Weakly Supervised Contrastive Learning Based on Semantic Grouping.}
As shown in Figure~\ref{fig:wsg}, we use semantic groupings derived from SAM to construct weakly supervised contrastive learning to optimize the description space of local features.
Our motivation is very intuitive: \textit{i.e.}, pixels belonging to the same semantic grouping should be closer in the description space, and on the contrary pixels of different groupings should be kept at a distance in the description space.
However, since two pixels belonging to the same grouping do not imply that their descriptors are the closest pair, forcing them to align will impair the discriminative properties of pixels within the same grouping. Therefore, semantic grouping can only provide weakly supervised constraints, and we maintain the discriminatory nature within the semantic grouping by setting a margin in optimization.
Given the sampling points set $P \in \mathbb{R}^{N}$, the positive sample average distance $D_{pos}$ can be defined as:
\begin{equation}
D_{pos}= \frac{1}{J} \sum^{J}_{i,j} {\rm dis} (P_{i},P_{j}), where~G(i)=G(j)~and~ i\ne j.
\end{equation}


Here ${\rm dis(P_{i},P_{j})}$ means calculate the Euclidean distance between the local descriptors corresponding to the two sampling points $P_{i}$ and $P_{j}$.
$G(\cdot)$ denotes the indexed semantic grouping category. $J$
denotes the number of positive samples, noting that since $J$ is not consistent for each image, we take the average to denote the positive sample distance.
Similarly, the negative sample average distance $D_{neg}$ can be defined as:

\begin{equation}
\begin{aligned}
D_{\text{neg}} &= \frac{1}{K} \sum^{K}_{i,j} \operatorname{dis}(P_{i},P_{j}), \\
&\quad \text{where } G(i) \neq G(j).
\end{aligned}
\end{equation}

\noindent where $K$ denotes the number of negative samples.
Thus, the final $\mathcal{L}_{wsc}$ loss can be defined as:
\begin{equation}
    \mathcal{L}_{wsc} = -\log(\frac{{\rm exp}(\max( D_{pos}, {\rm M}) / {\rm T})}{{\rm exp}(\max( D_{pos}, {\rm M}) + D_{neg}) / {\rm T)}}),
    \label{contrastive loss}
\end{equation}

where ${\rm M}$ is a margin parameter used to protect distinctiveness within semantic groupings, and ${\rm T}$ means the temperature coefficient.


\noindent\textbf{Edge Attention Guidance.} Edge regions are more worthy of the network's attention than mundane regions. \label{Edge Attention Guidance.}
On one hand, corner and edge points in the edge region are more likely to be detected as keypoints. On the other hand, the edge region contains rich information about the geometric structure thus contributing more to the discriminative nature of the local descriptor.
To enable the network to better capture the details of edge areas and improve the robustness of descriptors, we propose the Edge Attention Guidance Module, which can guide the network to focus on edge regions.
As shown in Figure~\ref{fig:method}, we first set up an edge head to predict the edge map ${E}^{'}$ and use the SAM output of the edge map for supervision.
 The edge loss $\mathcal{L}_{edge}$ is denoted as:
 \begin{equation}
     \mathcal{L}_{edge} = \sum_{i}^{H\times W} |E_{i} - E^{'}_{i}|.
 \end{equation}
We then fuse the predicted edge map ${E}^{'}$ into the local feature detection and description pipeline to bootstrap the network. 

Figure~\ref{fig:Edge_Learning} visualize the learning outcome of object boundaries under the guidance of SAM. With our EAG, SAMFeat learns accurate boundaries and edges effectively and efficiently. This prior knowledge of boundaries and edges will then aid feature detection and description.

\begin{figure*}[t] 
\begin{center}
  \includegraphics[width=0.9 \linewidth]{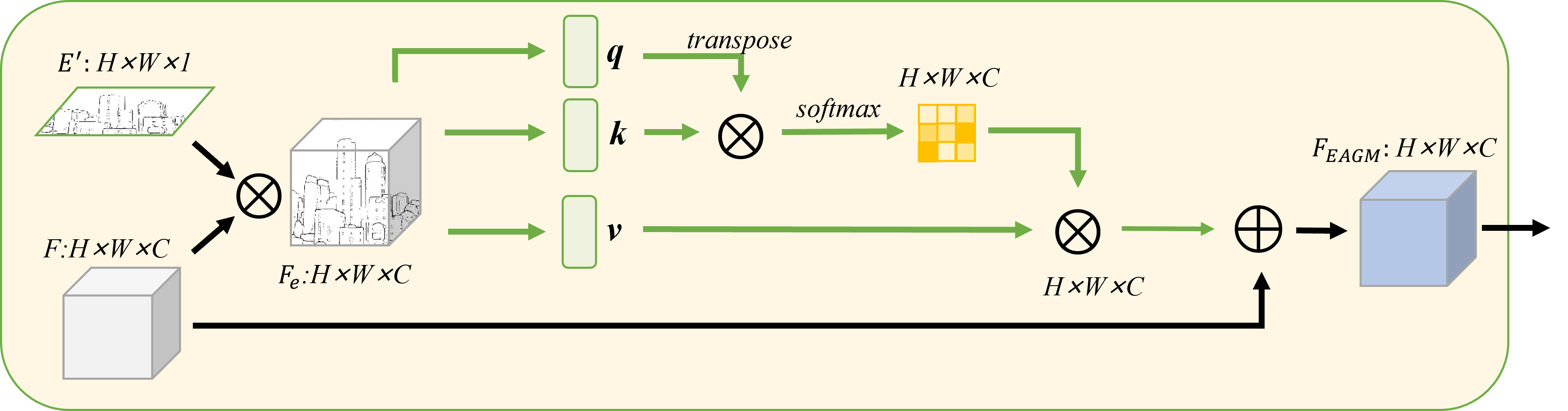}
\end{center}
   \caption{Details of Edge Attention Guidance Module.}
\label{fig:eagm}
\end{figure*}

\subsection{Guidance from SAM}
\label{Guidance from SAM}
\noindent \textbf{1) Learning Gifts From SAM.}:
As described in section~\ref{sec:3.2}, The carefully designed architecture enables SAMFeat to output the learned knowledge under the supervision of three designed loss modules. Therefore, we could summarize the guidance loss supervision from SAM $\mathcal{L}_{g}$ as:

\begin{equation}
 \mathcal{L}_{g} =  \mathcal{L}_{dis}+ \mathcal{L}_{edge}+ \mathcal{L}_{wsc}.
\end{equation}

With accurate loss supervision, SAMFeat is able to utilize the learned knowledge in later modules discussed in section~\ref{SAMFeat} to further aid feature learning and matching tasks.

\noindent \textbf{2) Guided Local Feature Detection}:
To aid feature detection in SAMFeat, the predicted edge map $E'$ from the edge head is performed with a pixel-wise dot product with the middle-level encoded feature representation $C_3$, shown in Figure~\ref{fig:method}. The product is added to $C_3$ for a residual purpose to obtain an edge-enhanced feature $C_3$. This feature will be used to generate a heatmap via the detection head to provide better local feature detection.

\noindent \textbf{3) Guided Local Feature Description}:
We filter the edge features by the predicted edge map and model the features of the edge region by a self-attention mechanism to encourage the network to capture the information of the edge region.
Specifically, the predicted edge map $E'$ from the edge head, and the multi-scale feature maps $F_{in}$ extracted from the backbone are fed into the Edge Attention Guidance Module. As shown in Figure~\ref{fig:eagm}, we first fuse $E'$ and $F_{in}$ by applying a pixel-wise dot product to obtain an edge-oriented feature map ${F}_{edge}$. Then we apply different convolutional transformations to the given ${F}_{edge}$ to get query $q$, key $k$, and value $v$ respectively. We then calculate the attention score using the dot product between query and key. Next, we use the \textit{softmax} function on the attention score to obtain the attention weight, which is used to calculate the edge-enhanced feature maps with the value feature vector. Finally, the edge-enhanced feature maps and the $F_{in}$ are added to obtain the output feature maps $F_{out}$.

\noindent\textbf{Total Loss.}
The total loss $\mathcal{L}$ can be defined as:
 \begin{equation}
     \mathcal{L} = \mathcal{L}_{g} + \mathcal{L}_{det} + \mathcal{L}_{des}
 \end{equation}
$\mathcal{L}_{g}$ is the loss function guided by SAM's knowledge defined in section~\ref{Guidance from SAM}, while $\mathcal{L}_{det}$ is the cross entropy loss for supervised keypoint detection and $\mathcal{L}_{des}$ is the attention weighted triplet loss from MTLDesc~\cite{wang2022mtldesc} for optimizing the local descriptors. Individual weights for each loss are not assigned: each loss shares equal weights. This independence from hyper-parameters, again, shows the robustness of SAMFeat.

\section{Experiments}

\subsection{Implementation.} To generate our training data with dense pixel-wise correspondences, we rely on the MegaDepth dataset \cite{li2018megadepth}, a rich resource containing image pairs with known pose and depth information from 196 diverse scenes. Specifically, we use MTLDesc~\cite{wang2022mtldesc}~\footnote{https://github.com/vignywang/MTLDesc} released megedepth image and the correspondence ground truth for training. In our experiment, we meticulously configured the parameters to establish a consistent and efficient training process. Hyper-parameters are set as follows. The learning rate of 0.001 enables gradual parameter updates, and the weight decay of 0.0001 helps control model complexity and mitigate overfitting. With a batch size of 14, our model processes 14 samples per iteration, striking a balance between computational efficiency and convergence. $\rm M$ and $\rm T$ are set to 0.07 and 5. Training spans 30 epochs to ensure comprehensive exposure to the data, with a total training time of 3.5 hours. By meticulously defining these parameters and configurations, we establish a robust experimental setup that ensures replicability and accurate evaluation of our model's performance. More detailed information about parameter tuning and ablation experiments can be found in the supplementary material.

\begin{figure*}
\begin{center}
  \includegraphics[width=1 \linewidth]{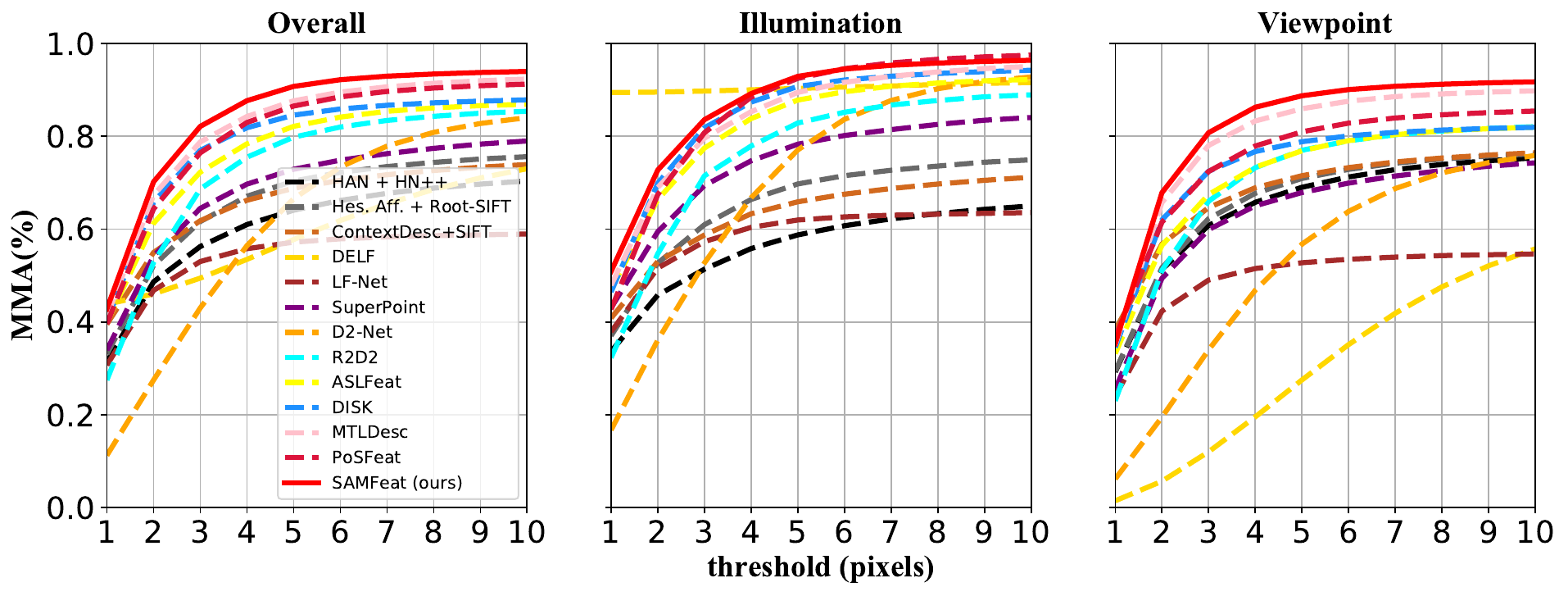}
\end{center}
   \caption{
   Comparisons on HPatches dataset with different thresholds Mean Matching Accuracy. Our SAMFeat achieves higher average local feature matching accuracy than other state-of-the-art methods at all thresholds.}
\label{fig: HPatches Curve}
\end{figure*}

\subsection{Image Matching.} We evaluate the performance of our method in the image-matching tasks on the most popular feature learning and matching benchmark: HPatches \cite{hpatches}. The HPatches dataset consists of 116 sequences of image patches extracted from a diverse range of scenes and objects. Each image patch is associated with ground truth annotations, including key point locations, descriptors, and corresponding homographies. We follow the same evaluation protocol as in D2-Net \cite{dusmanu2019d2}, where eight unreliable scenes are excluded. To ensure an equitable comparison, we align the features extracted by each method through nearest-neighbor matching. A match is deemed accurate if its estimated reprojection error is lower than a predetermined matching threshold. The threshold is systematically varied from 1 to 10 pixels, and the mean matching accuracy (MMA) across all pairs is recorded, indicating the proportion of correct matches relative to potential matches. Subsequently, the area under the curve (AUC) is computed at 5px based on the MMA. The comparison between SAMFeat and other state-of-the-art methods on HPatches image matching is visualized in Figure \ref{fig: HPatches Curve}. The MMA @3 threshold against other state-of-the-art methods under each threshold is listed in  Table \ref{HPatches comp}. SAMFeat achieved the highest MMA @3 even compared to the most updated feature learning model in 2023 top-tier conferences. 

\begin{table}[h]
\centering
\caption{Image Matching Performance Comparison on HPatches dataset.}
\setlength{\tabcolsep}{3mm}{
\renewcommand\arraystretch{1.5} {
\begin{tabular}{l|c|c}
\toprule[1.5px]
\textbf{Methods} & \textbf{MMA @3} & \textbf{AUC @5} \\
\hline
\hline
SIFT~$_{IJCV' 2012}$ \cite{lindeberg2012scale} & 50.1 & 49.6 \\
HardNet~$_{NeurIPS' 2017}$ \cite{mishchuk2017working} & 62.1 &  56.9 \\
DELF~$_{ICCV' 2017}$ \cite{noh2017large} & 50.7 & 49.7 \\ 
SuperPoint~$_{CVPRW' 2018}$ \cite{detone2018superpoint} & 65.7 & 59.0\\
Lf-net~$_{NeurIPS' 2018}$ \cite{ono2018lf} & 53.2 & 48.7 \\ 
ContextDesc~$_{CVPR' 2019}$ \cite{luo2019contextdesc} & 63.2 & 58.3 \\ 
D2Net~$_{CVPR' 2019}$ \cite{dusmanu2019d2} & 40.3 & 37.8\\
R2D2~$_{NeurIPS' 2019}$ \cite{revaud2019r2d2} & 72.1 &  64.1\\
DISK~$_{NeurIPS' 2020}$ \cite{tyszkiewicz2020disk} & 72.2 & 69.8\\
ASLFeat~$_{CVPR' 2020}$ \cite{luo2020aslfeat} &  72.2 &  66.9 \\ 
LLF~$_{WACV' 2021}$ \cite{suwanwimolkul2021learning} &  74.0 & 66.8 \\ 
Key.Net~$_{TPAMI' 2022}$ \cite{barroso2022key} & 72.1 & 56.0 \\
ALIKE~$_{TMM' 2022}$ \cite{zhao2022alike} &  70.5 & 69.0 \\ 
MTLDesc~$_{AAAI' 2022}$ \cite{wang2022mtldesc} & 78.7 & 71.4\\
PoSFeat~$_{CVPR' 2022}$ \cite{li2022decoupling} &  75.3 & 69.2 \\ 
SFD2~$_{CVPR' 2023}$ \cite{xue2023sfd2} & 70.6 & 64.8 \\
TPR~$_{CVPR' 2023}$ \cite{wang2023learning} & 79.8 & 73.0\\
\textbf{SAMFeat (Ours)} & \textbf{82.2} & \textbf{74.4}\\
\bottomrule[1.5px] 
\end{tabular}}}
\label{HPatches comp}
\end{table}

\subsection{Visual Localization.} To further validate the efficacy of our approach when dealing with intricate tasks, we assess its performance in the area of visual localization. This task involves estimating the camera's position within a scene using an image sequence and serves as an evaluation benchmark for local feature performance in long-term localization scenarios, without requiring a dedicated localization pipeline. We utilize the Aachen Day-Night v1.1 dataset~\cite{sattler2018benchmarking} to showcase the impact on visual localization tasks. To ensure fairness in the assessment, we employ a predefined visual localization pipeline\footnote{https://github.com/GrumpyZhou/image-matching-toolbox} based on colmap provided by benchmark\footnote{https://www.visuallocalization.net}. This pipeline operates as follows: Initially, custom features extracted from the database's images are employed to construct a structure-from-motion model. Subsequently, the query images are registered within this model using the same custom features. For keypoints matching, we utilize the mutual nearest neighbor approach to effectively filter out outliers. And we set the number of features in this experiment to $10000$.
We tally the number of accurately localized images under three distinct error thresholds, namely (0.25m, 2°), (0.5m, 5°), and (5m, 10°), signifying the maximum allowable position error in both meters and degrees. Referring to Table~\ref{Loc comp}, we categorize current state-of-the-art methods into two categories: $G$ contains methods that are designed for general feature learning tasks; $L$ contains methods that are designed, tuned, and tested specifically for localization tasks, and they typically perform poorly outside of specific localization scenarios, as shown in Table~\ref{HPatches comp}. SAMFeat achieved the top performance among all general methods, while also revealing a competitive performance among methods that are designed specifically for visualization.

\begin{table}[t]
\centering
\caption{Visual Localization Performance Comparison on Aachen V1.1. Category ``L'' means local feature methods specifically designed for visual localization tasks, and ``G'' means generalized local feature methods.}
\setlength{\tabcolsep}{1.5mm}\scalebox{0.9}{
\renewcommand\arraystretch{1.2} {
\begin{tabular}{clcc}
\toprule[1.5px]
\multicolumn{1}{c}{\multirow{3}{*}{Category}} & \multicolumn{1}{c}{\multirow{3}{*}{Method}} & \multicolumn{2}{c}{Accuracy @ Thresholds (\%) ↑}        \\ \cline{3-4} 
\multicolumn{1}{c}{}                          & \multicolumn{1}{c}{}                        & \multicolumn{1}{c}{Day}    & \multicolumn{1}{c}{Night} \\ \cline{3-4} 
\multicolumn{1}{c}{}                          & \multicolumn{1}{c}{}                        & \multicolumn{2}{c}{0.25m,2°/0.5m,5°/5m,10°}             \\ \hline \hline
\multirow{2}{*}{L}                              & SeLF~$_{TIP' 22}$    \cite{fan2022learning}                                      & $-$         & 75.0 / 86.8 / 97.6         \\
                                                & SFD2~$_{CVPR' 2023}$     \cite{xue2023sfd2}                                    & 88.2 / 96.0 / 98.7          & 78.0 / 92.1 / 99.5         \\  \hline
\multirow{10}{*}{G}                             & SIFT~$_{IJCV' 12}$ \cite{sift}                                         & 72.2 / 78.4 / 81.7          & 19.4 / 23.0 / 27.2         \\
& SuperPoint~$_{CVPRW' 18}$ \cite{superpoint}                                  & 87.9 / 93.6 / 96.8          & 70.2 / 84.8 / 93.7         \\
& D2-Net~$_{CVPR' 19}$   \cite{d2net}                                    & 84.1 / 91.0 / 95.5          & 63.4 / 83.8 / 92.1         \\
& R2D2~$_{NeurIPS' 19}$     \cite{r2d2}                                    & 88.8 / 95.3 / 97.8          & 72.3 / 88.5 / 94.2         \\
& ASLFeat~$_{CVPR' 20}$  \cite{aslfeat}                                    & 88.0 / 95.4 / 98.2          & 70.7 / 84.3 / 94.2         \\
& CAPS~$_{ECCV' 20}$  \cite{wang2020caps}                                  & 82.4 / 91.3 / 95.9          & 61.3 / 83.8 / 95.3         \\
 & LISRD~$_{ECCV' 20}$  \cite{pautrat2020online}                                  & $-$           & 73.3/ 86.9 / 97.9         \\
& DISK~$_{NeurIPS' 22}$ \cite{tyszkiewicz2020disk}                                  & $-$                            & 73.8 / 86.2 / 97.4         \\
& PoSFeat~$_{CVPR' 22}$ \cite{li2022decoupling}                                  & $-$                         & 73.8 / 87.4 / \textbf{98.4}         \\
& MTLDesc~$_{AAAI' 22}$ \cite{wang2022mtldesc}                                     & $-$                           & 74.3  / 86.9 / 96.9         \\
& TR~$_{CVPR' 23}$   \cite{wang2023learning}                                        & $-$                            & 74.3  / 89.0 /  \textbf{98.4}         \\
& SAMFeat (Ours)                      & \textbf{90.2} / \textbf{96.0} / \textbf{98.5} & \textbf{75.9}  / \textbf{89.5} / 95.8        \\ 
                                                \bottomrule[1.5px] 
\end{tabular}
}}
\label{Loc comp}
\end{table}

\subsection{3D Reconstruction.}
\begin{table}[h]
\caption{Evaluation on ETH 3D reconstruction benchmark \cite{schonberger2017comparative}. The top two results
are marked with \textbf{bold} (1st) and \underline{underline} (2nd).}\scalebox{0.8}{
\begin{tabular}{ccccccc}
\hline
\multicolumn{7}{c}{ETH benchmark} \\ \hline
\multicolumn{1}{c|}{Datasets} & \multicolumn{1}{c|}{Methods} & \begin{tabular}[c]{@{}c@{}}\#Reg.\\ Images\end{tabular} & \begin{tabular}[c]{@{}c@{}}\#Sparse\\ Points\end{tabular} & \begin{tabular}[c]{@{}c@{}}Track\\ Length\end{tabular} & \begin{tabular}[c]{@{}c@{}}Reproj.\\ Error\end{tabular} & \begin{tabular}[c]{@{}c@{}}\#Dense\\ Points\end{tabular} \\ \hline
\multicolumn{1}{c|}{\multirow{6}{*}{\begin{tabular}[c]{@{}c@{}}Madrid \\ Metropolis \\ 1344 images\end{tabular}}} & \multicolumn{1}{c|}{SuperPoint \cite{superpoint}} & 438 & 29K & 9.03 & 1.02px & 1.55M \\
\multicolumn{1}{c|}{} & \multicolumn{1}{c|}{D2-Net \cite{d2net}} & 495 & 144k & 6.39 & 1.35px & 1.46M \\
\multicolumn{1}{c|}{} & \multicolumn{1}{c|}{ASLFeat \cite{aslfeat}} & 613 & 96k & 8.76 & 0.90px & \textbf{2.00M} \\
\multicolumn{1}{c|}{} & \multicolumn{1}{c|}{DISK \cite{tyszkiewicz2020disk}} & 677 & 213K & 7.89 & 1.14px & 1.87M \\
\multicolumn{1}{c|}{} & \multicolumn{1}{c|}{AWDesc-CA \cite{wang2023attention}} & \underline{864} & \underline{278K} & \underline{9.52} & \textbf{0.96px} & 1.65M \\
\multicolumn{1}{c|}{} & \multicolumn{1}{c|}{SAMFeat} & \textbf{892} & \textbf{282K} & \textbf{9.84} & \underline{0.93px} &  \underline{1.90M} \\ \hline
\multicolumn{1}{c|}{\multirow{6}{*}{\begin{tabular}[c]{@{}c@{}}Gendar- \\
menmarkt\\ 1463 images\end{tabular}}} & \multicolumn{1}{c|}{SuperPoint \cite{superpoint}} & 967 & 93k & \underline{7.22} & 1.03px & 3.81M \\
\multicolumn{1}{c|}{} & \multicolumn{1}{c|}{D2-Net \cite{d2net}} & 965 & 310K & 5.55 & 1.28px & 3.15M \\
\multicolumn{1}{c|}{} & \multicolumn{1}{c|}{ASLFeat \cite{aslfeat}} & 1040 & 221K & \textbf{8.72} & 1.00px & \textbf{4.01M} \\
\multicolumn{1}{c|}{} & \multicolumn{1}{c|}{DISK \cite{tyszkiewicz2020disk}} & 1218 & \underline{588K} & 6.02 & 0.98px & 3.62M \\
\multicolumn{1}{c|}{} & \multicolumn{1}{c|}{AWDesc-CA \cite{wang2023attention}} & \underline{1354} & 548K & 6.94 & \underline{0.95px} & 3.86M \\
\multicolumn{1}{c|}{} & \multicolumn{1}{c|}{SAMFeat} & \textbf{1370} & \textbf{596K} & 7.02 & \textbf{0.93px} & \underline{3.91M} \\ \hline
\multicolumn{1}{c|}{\multirow{6}{*}{\begin{tabular}[c]{@{}c@{}}Tower \\ of \\ London \\ 1576 images\end{tabular}}} & \multicolumn{1}{c|}{SuperPoint \cite{superpoint}} & 681 & 52K & 8.76 & 0.96px & 2.77M \\
\multicolumn{1}{c|}{} & \multicolumn{1}{c|}{D2-Net \cite{d2net}} & 708 & 287K & 5.20 & 1.34px & 2.86M \\
\multicolumn{1}{c|}{} & \multicolumn{1}{c|}{ASLFeat \cite{aslfeat}} & 821 & 222K & 12.52 & 0.92px & \textbf{3.06M} \\
\multicolumn{1}{c|}{} & \multicolumn{1}{c|}{DISK \cite{tyszkiewicz2020disk}} & 985 & 517K & 5.90 & 1.02px & \underline{3.00M} \\
\multicolumn{1}{c|}{} & \multicolumn{1}{c|}{AWDesc-CA \cite{wang2023attention}} & \underline{1414} & \underline{563K} & \underline{12.88} & \underline{0.88px} & 2.89M \\
\multicolumn{1}{c|}{} & \multicolumn{1}{c|}{SAMFeat} & \textbf{1443} & \textbf{587K} & \textbf{13.01} & \textbf{0.84px} & 2.91M \\ \hline
\end{tabular}}
\label{3DReconsturction}
\end{table}

We utilize the ETH benchmark \cite{schonberger2017comparative} to evaluate the performance of our method on the 3D reconstruction task. Three medium-scale datasets from the benchmark are used for this purpose. We perform exhaustive image matching on all three collections and apply ratio test filtering with a default threshold of 0.8 to eliminate incorrect matches. The reconstruction protocol follows the COLMAP \cite{colmap} pipeline, which involves running Structure-from-Motion (SfM) first and then Multi-View Stereo (MVS) to generate the dense point cloud models. As shown in Table \ref{3DReconsturction}, SAMFeat consistently produces the highest number of registered images and sparse points, along with competitive track length and reprojection error, demonstrating its robustness and accuracy in 3D reconstruction tasks.


\subsection{Ablation Study.} We conduct ablation studies on different aspects to support our claim and illustrate the necessity of each of our contributed modules.

\noindent\textbf{Ablation on Designed Modules.} Table~\ref{Detailed Ablation} demonstrates the efficacy of the components within our network as we progressively incorporate Attention-weighted Semantic Relation Distillation (ASRD), Weakly Supervised Contrastive Learning Based on Semantic Grouping (WCS), and Edge Attention Guidance (EAG). The effectiveness of each component is reflected by the Mean Matching Accuracy at the pixel three threshold on the HPatches Image Matching task. Our baseline is trained using SuperPoint~\cite{detone2018superpoint} structure along with its detector supervision and attention-weighted triplet loss~\cite{wang2022mtldesc} for descriptor learning. Following the addition of the ASRD, the model's performance notably improves due to better image feature learning. The introduction of the WCS further enhances accuracy by augmenting the discriminative power of descriptors with semantics. It demonstrates superior performance as it better preserves the inner diversity of objects by optimizing sample ranks. Lastly, the inclusion of the EAG bolsters the network's capability to embed object edge and boundary information, resulting in further enhancements in accuracy.

\begin{table}[h]
\centering
\caption{Detailed Ablation Study on SAMFeat. \checkmark means denotes applied components. The results of MMA@ 3 on HPatches of removing each component individually in addition to applying the components sequentially are reported.}
\setlength{\tabcolsep}{5mm}{
\renewcommand\arraystretch{1.2} {
\begin{tabular}{cccc}
\toprule[1.5px]
\multicolumn{1}{c|}{ASRD} & \multicolumn{1}{c|}{EAG} & \multicolumn{1}{c|}{WCS} & \multicolumn{1}{c}{MMA @3} \\ \hline \hline
                                  &                          &                        & 75.7              \\
\checkmark                        &                          &                        & 78.6               \\
\checkmark                        & \checkmark               &                         & 80.9              \\
\checkmark                        & \checkmark               & \checkmark              & \bf 82.2          \\
\checkmark                        &                          & \checkmark              & 81.2              \\
                                  & \checkmark               & \checkmark              & 79.4              \\
 \bottomrule[1.5px]
\end{tabular}
}}
\label{Detailed Ablation}
\end{table}

\noindent\textbf{Attention-weighted Semantic Relation Distillation Versus Direct Semantic Feature Distillation} 
Table~\ref{Ablation ASRD} highlights the performance differences between two approaches for distilling image features from the SAM encoder: our proposed Attention-weighted Semantic Relation Distillation (ASRD) and Direct Semantic Feature Distillation (DSFD). The effectiveness of these approaches is evaluated using the Mean Matching Accuracy at the pixel three threshold on the HPatches Image Matching task. The results demonstrate that while DSFD achieves a Mean Matching Accuracy (MMA) of 76.9, the ASRD method significantly enhances performance, achieving an MMA of 78.6. This indicates that ASRD provides superior feature learning by effectively capturing and distilling the semantic relationships within the image features.

\begin{table}[h]
\centering
\caption{Ablation test on the Attention-weighted Semantic Relation Distillation (ASRD).}

\setlength{\tabcolsep}{8mm}{
\renewcommand\arraystretch{1.5} {
\begin{tabular}{c|c}
\toprule[1.5px]
Module Selected         & MMA @3 \\ \hline
DSFD       & 76.9     \\ \hline
ASRD & \bf 78.6     \\ \bottomrule[1.5px]
\end{tabular}}}
\label{Ablation ASRD}
\end{table}

\noindent\textbf{Hyper-Parameters in Weakly Supervised Contrastive Module.}
Table~\ref{Ablation WCS} presents the impact of different hyper-parameter values on the performance of Weakly Supervised Contrastive Learning Based on Semantic Grouping (WCS). Specifically, the margin parameter $M$ is used to maintain distinctiveness within semantic groupings, while $T$ represents the temperature coefficient. The effectiveness of these hyper-parameters is evaluated using the Mean Matching Accuracy at the pixel three threshold on the HPatches Image Matching task. The results demonstrate that varying $M$ and $T$ values impact performance. Notably, the combination of $M = 0.07$ and $T = 5$ achieves the highest accuracy, with an MMA of 82.2. This configuration provides an optimal balance, enhancing the discriminative power of the descriptors by effectively preserving the inner diversity of objects.

\begin{table}[h]
\centering
\caption{Ablation test on hyper-parameters on WCS.}
\setlength{\tabcolsep}{10mm}{
\renewcommand\arraystretch{1.5} {
\begin{tabular}{c|c}
\toprule[1.5px]
(M, T) & MMA @3 \\ \hline \hline
0, 1      & 80.9   \\
0.03, 1   & 81.2   \\
0.05, 1   & 80.3   \\
\bf 0.07, 1   & \bf 81.3   \\
0.09, 1   & 80.8   \\
0.11, 1   & 80.5   \\ \hline
0.07, 3   & 81.6   \\
\bf 0.07, 5   & \bf 82.2   \\
0.07, 7   & 82.0   \\
0.07, 9   & 81.8    \\ \bottomrule[1.5px]
\end{tabular}}}
\label{Ablation WCS}
\end{table}

\noindent\textbf{Training Samples.} Table~\ref{training samples} provides a comparative analysis of various methods based on the number of training samples used and their Mean Matching Accuracy (MMA) at the pixel three threshold on the HPatches dataset. Despite being trained on a relatively small dataset of 23,600 images, SAMFeat achieves an outstanding MMA@3 of 82.2, surpassing all other methods. This result underscores the efficiency and robustness of SAMFeat in achieving superior performance with limited training data. For instance, SuperPoint, trained on 80,000 images, achieves an MMA@3 of 64.5, while ASLFeat, utilizing a significantly larger dataset of 1,600,000 images, achieves an MMA@3 of 72.3. Similarly, methods like D2Net and R2D2, despite having access to larger or comparable training datasets, attain lower MMA@3 scores of 42.9 and 68.6, respectively. These comparisons highlight the effectiveness of SAMFeat's design in leveraging a modest amount of training data to achieve top-tier performance.

\begin{table}[h]
\caption{Comparisons on the number of Training Samples.}
\renewcommand\arraystretch{1.5}{
\resizebox{\linewidth}{!}{
\begin{tabular}{c|c|c|c}
\toprule[1.5px]
Method     & Source                                                          & Images          & MMA@3          \\ \hline\hline
SuperPoint & COCO                                                            & 80,000          & 64.5          \\ 
D2Net      & MegaDepth                                                       & 617,774         & 42.9          \\ 
R2D2       & Aachen and Web images                                           & \textbf{12,083} & 68.6          \\ 
ASLFeat    & GL3D                                                            & 1600,000        & 72.3          \\ 
MTLDesc    & MegaDepth                                                       & 23,600          & 78.7 
   \\ 
SFD2    & Aachen and Web images                                              & \textbf{12,083} & 70.6 
   \\ 
TRR    & COCO +  Image Matching Challenge                                    & 106,000         & 79.8 
   \\ 
SAMFeat    & MegaDepth                                                       & 23,600          & \textbf{82.2} \\  \bottomrule[1.5px]
\end{tabular}
}}
\label{training samples}
\end{table}

\noindent\textbf{Training Time}
Table~\ref{Training Time} presents a quantitative analysis of the overhead training time costs associated with incorporating additional loss functions into our method. The table details the incremental training time required for each combination of the Attention-weighted Semantic Relation Distillation (ASRD), Edge Attention Guidance (EAG), and Weakly Supervised Contrastive Learning Based on Semantic Grouping (WCS) loss components. Note that our method only requires training for 6 hours using two Nvidia RTX 3090 GPUs. Compared to other work like ASLFeat (42 hours on a single NVIDIA RTX 2080Ti) and TRR (30 hours for training with two NVIDIA-A100 GPUs), this demonstrates a totally reproducible cost for individual researchers.

Even though each additional loss function introduces some extra training time, the tradeoff is justified by the minimal increase in time and the corresponding improvement in accuracy. The final training time remains acceptable, demonstrating the lightweight nature of our approach. This efficiency makes our method easily implementable and resource-efficient, highlighting its reproducibility and practicality for individual researchers in the field of feature learning and description.

\begin{table}[h]
\centering
\caption{A quantitative analysis on the overhead training time costs for adding the extra loss functions. \checkmark means denotes applied loss components.}
\setlength{\tabcolsep}{5mm}{
\renewcommand\arraystretch{1.2} {
\begin{tabular}{cccc}
\toprule[1.5px]
\multicolumn{1}{c|}{ASRD} & \multicolumn{1}{c|}{EAG} & \multicolumn{1}{c|}{WCS} & \multicolumn{1}{c}{Training time in Hours} \\ \hline \hline
                          &                          &                        & 3.6                         \\
\checkmark                        &                          &                         & 4.7                         \\
\checkmark                        & \checkmark                         &                         & 5.1                         \\
\checkmark                        & \checkmark                         & \checkmark                        & 6.0          
\\ \bottomrule[1.5px]
\end{tabular}
}}
\label{Training Time}
\end{table}

\noindent\textbf{Ablation on Loss Weights}
Table~\ref{Ablation Loss Weight} presents the results of an ablation test that explores the impact of adjusting the loss weight of Edge Attention Guidance (EAG) without incorporating Weakly Supervised Contrastive Learning Based on Semantic Grouping (WCS). The effectiveness of these adjustments is measured using the Mean Matching Accuracy at the pixel three threshold on the HPatches dataset. The results indicate that merely adjusting the loss weights of EAG does not achieve the same effectiveness as incorporating WCS. Specifically, with a fixed ASRD weight of 1.0, varying the EAG weight from 0.5 to 1.5 yields incremental improvements in MMA@3, peaking at 81.0. However, this peak still falls short of the performance enhancements observed when WCS is included, highlighting the unique contribution of WCS to the overall accuracy.

\begin{table}[h]
\centering
\caption{Ablation test on adjusting the loss weight of EAG without WCS. The MMA @3 on HPatches are recorded, showing that it is difficult to achieve the effect of imposing WCS by only adjusting the loss weights.}
\setlength{\tabcolsep}{6mm}{
\renewcommand\arraystretch{1.5}{
\begin{tabular}{c|c|c}

\toprule[1.5px]
Weights of ASRD & Weights of EAG  & MMA @3 \\ \hline \hline
1.0 & 0.5      & 80.7     \\ 
1.0 & 1.0 & 80.9     \\ 
1.0 & 1.5 & 81.0     \\ 
\hline
\end{tabular}}}
\label{Ablation Loss Weight}
\end{table}

\subsection{Inference Speed.} To further demonstrate SAMFeat's high efficiency and fast inference speed, we conduct a comparison of inference time between other state-of-the-art methods in table \ref{speed}.
We assessed the running speed of various methods using open-source code. In Table \ref{speed}, our approach demonstrated exceptionally competitive performance while maintaining a fast inference speed among many lightweight methods.

\begin{table}[h]
\centering
\caption{Comparisons on the Inference Speed. The speed is calculated as the average feature extraction inference speed on HPatches (480 × 640) with the same setting}
\setlength{\tabcolsep}{1mm}{
\renewcommand\arraystretch{1.5} {
\begin{tabular}{ccccccc}
\hline
Methods         & Superpoint & D2-Net & SFD2  & MTLDesc & R2D2 & SAMFeat \\ \hline
Inference Speed & 31FPS      & 6FPS   & 11FPS & 24FPS   & 8FPS & 21FPS  \\ \hline
\end{tabular}}}
\label{speed}
\end{table}

\subsection{Matching Visualization.}
\begin{figure}
\captionsetup{justification=centering}
\centering
  \includegraphics[width=1 \linewidth]{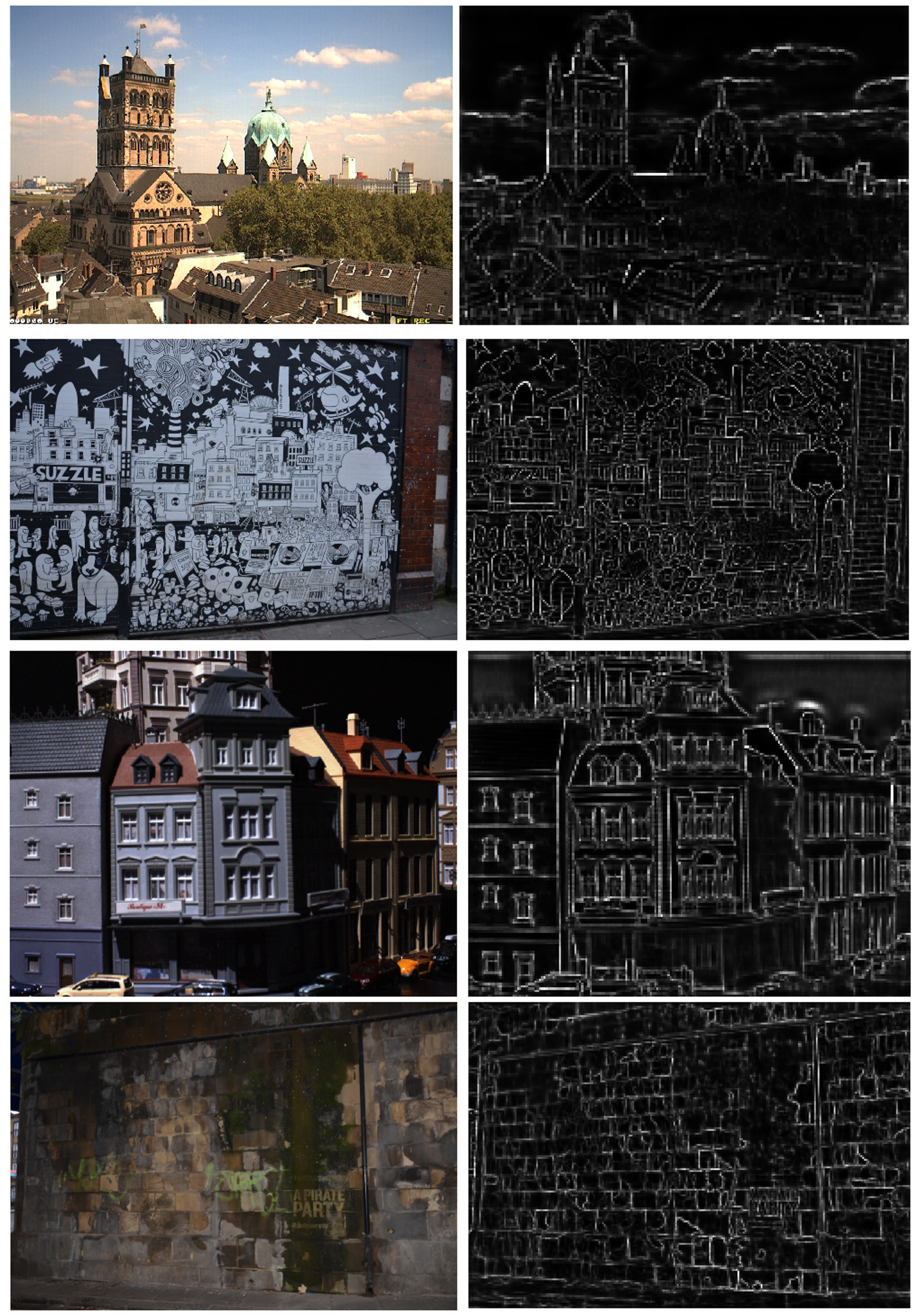}
   \caption{Left: Random images selected from HPatches. Right: Learned edge boundaries from SAMFeat under the guide of SAM.}

\label{fig:Edge_Learning}
\end{figure}

As mentioned in Section \ref{sec:3.2} in the full paper, SAMFeat learns edge maps from SAM and utilizes Edge Attention Guidance (EAG) to further enhance the precision of local feature detection and description by encouraging the network to prioritize attention to the edge region. Figure~\ref{fig:Edge_Learning} demonstrates the learning outcome of SAMFeat. With the fine-grained object boundaries from SAM, SAMFeat is able to learn clear object edges. This illustrates two things: first, the encoded feature that is used to generate the edge map contains rich edge information, and second, with a clear and accurate generated edge map and EAG, SAMFeat is able to better capture the details of edge areas and improve the robustness of local descriptors.


\section{Limitations}
Although other visual foundation models like DINO \cite{oquab2023dinov2} or SEEM \cite{zou2024segment} could potentially serve as alternative teachers, the focus of our study was specifically on SAM. Our methodology was designed around the unique capabilities of SAM, and therefore, further investigation into alternative teachers was not pursued in this study. Future research could explore the applicability and potential advantages of employing other visual foundation models as teachers for local feature learning tasks.

\section{Conclusion}
In this study, We introduce SAMFeat, a local feature learning method that harnesses the power of the Segment Anything Model (SAM). SAMFeat encompasses three innovations. Firstly, we introduce Attention-weighted Semantic Relation Distillation (ASRD), an auxiliary task aimed at distilling the category-agnostic semantic information acquired by the SAM encoder into the local feature learning network. Secondly, we present Weakly Supervised Contrastive Learning Based on Semantic Grouping (WSC), a technique that leverages the semantic groupings derived from SAM as weakly supervised signals to optimize the metric space of local descriptors. Furthermore, we engineer the Edge Attention Guidance (EAG) mechanism to elevate the accuracy of local feature detection and description. Our comprehensive evaluation of tasks such as image matching on HPatches and long-term visual localization on Aachen Day-Night consistently underscores the remarkable performance of SAMFeat, surpassing previous methods.


\normalem
\bibliographystyle{IEEEtran}
\bibliography{ref}

\end{document}